\documentclass[letterpaper, 10pt, conference]{ieeeconf}  

\IEEEoverridecommandlockouts                              
\usepackage{graphicx}
\usepackage{amsmath}
\usepackage{amssymb}
\usepackage{mathtools}
\usepackage{booktabs}
\usepackage{tabularx}
\usepackage{url}
\usepackage{color}
\usepackage{caption}
\usepackage{subcaption}
\graphicspath{{figs/}}

\newcommand{\Kb}{\mathcal{K}}

\title{\LARGE \bf
AURORA: A Natural Language--Driven Agentic Framework \\
for Understanding, Reasoning, and Orchestrating Reliable \\
Air--Ground Co-Simulation
}

\author{Keshu Wu$^{1}$, Hao Zhang$^{1}$, Rui Gan$^{2}$, Xiangbo Gao$^{3}$, Xiaopeng Li$^{2}$, Zhengzhong Tu$^{3}$, and Yang Zhou$^{1,\dagger}$
\thanks{$^{1}$Keshu Wu, Hao Zhang, and Yang Zhou are with the Zachry Department of Civil and Environmental Engineering, Texas A\&M University, College Station, TX, USA.
        {\tt\small \{keshuw, zhanghao230, yangzhou295\}@tamu.edu}}%
\thanks{$^{2}$Rui Gan and Xiaopeng Li are with the Department of Civil and Environmental Engineering, University of Wisconsin--Madison, Madison, WI, USA.
        {\tt\small \{rgan6, xli2485\}@wisc.edu}}%
\thanks{$^{3}$Xiangbo Gao and Zhengzhong Tu are with the Department of Computer Science and Engineering, Texas A\&M University, College Station, TX, USA.
        {\tt\small \{xiangbog, tzz\}@tamu.edu}}%
\thanks{$^{\dagger}$Corresponding author: Yang Zhou, {\tt\small yangzhou295@tamu.edu}}%
}

\begin{document}

\maketitle
\thispagestyle{plain}
\pagestyle{plain}


\begin{abstract}
Air--ground transportation research increasingly relies on co-simulation, yet constructing scenarios remains labor-intensive and difficult to validate. More importantly, a generated scenario may execute successfully while failing to realize the spatial, temporal, communication, or behavioral relationships requested by the user. This paper presents AURORA, a natural-language-driven agentic framework that treats air--ground scenario generation as a process of compilation with verification. Central to AURORA is the Air--Ground Scenario Graph (AGSG), a typed intermediate representation that explicitly connects agents, aerial missions, events, communication links, success conditions, and their cross-domain dependencies. This shared representation enables simulator-grounded parsing, joint road--airspace grounding, temporal planning, pre-execution feasibility checking, trace-based runtime verification, failure localization, and bounded repair within a unified workflow. We further introduce AURORA-Bench to evaluate not only whether generated scenarios execute, but whether they faithfully realize the requested interactions. Experiments across multiple language models show that structured execution substantially improves reliability, while runtime verification exposes silent failures that completion-based evaluation overlooks. Localized repair further resolves many violations without regenerating the entire scenario. The results show that reliable scenario generation requires verifying realized behavior, not merely executable code, and demonstrate the value of explicit intermediate representations for verifiable and repairable language-driven co-simulation.
\end{abstract}

\section{Introduction}
\label{sec:intro}

Autonomous driving, intelligent transportation systems, and the emerging low-altitude economy are creating mobility systems in which unmanned aerial vehicles (UAVs)~\cite{feng2024u2udata}, roadside infrastructure~\cite{yang2026edge}, connected vehicles~\cite{xu2022v2x,wu2025digital}, and other traffic participants~\cite{gao2025langcoop} must perceive, communicate, and coordinate within a shared environment. Air--ground simulation~\cite{gao2025airv2x,zeng2026carla,zhang2025virtual} provides a practical testbed for studying such interactions before deployment, particularly for applications such as cooperative perception, emergency response, and connected transportation. Yet constructing such experiments remains labor-intensive. Researchers must manually coordinate map selection, valid actor placement, UAV routing, simulator synchronization, event triggers, communication links, and success criteria, and these steps must be repeated for each new scenario. As a result, the scale and diversity of air--ground experiments are often limited not by simulator capability, but by the effort required to author and validate scenarios.

Natural-language and generative-AI interfaces offer a promising way to reduce this burden and improve interaction with connected and automated transportation systems~\cite{wang2025generative,wu2025v2x,gao2025automated}. Recent work has explored language-driven traffic scenario generation and editing~\cite{ruan2024traffic,zhang2024chatscene,chang2024llmscenario,sheng2025talk2traffic} and natural-language UAV control in simulation~\cite{phadke2024integrating,shibu2026skysim,koubaa2025agentic}. However, air--ground scenario generation requires more than translating a prompt into executable code. A scenario can run while violating requested spatial, temporal, perceptual, or communication relations---for example, through invalid placement, premature mission transitions, missed messages, or interactions that never occur. Such failures are difficult to detect when generated scripts lack an explicit specification of what should happen. Specification-based simulation analysis and safety evaluation provide foundations for detecting such failures~\cite{dreossi2019verifai,xu2022safebench}, but air--ground scenarios additionally require joint feasibility across road geometry, three-dimensional airspace, sensing, communication, and event timing. The challenge is therefore not only to generate executable scenarios, but to verify that the requested interactions are realized before and during execution.

We present \textbf{AURORA}\footnote{Project page: \url{https://keshuw95.github.io/AURORA/}} (\emph{A Natural Language--Driven \textbf{A}gentic Framework for \textbf{U}nderstanding, \textbf{R}easoning, and \textbf{O}rchestrating \textbf{R}eliable \textbf{A}ir--Ground Co-Simulation}), which treats language-driven scenario generation as \emph{compilation with verification}. Natural-language requests are converted into typed, provenance-aware specifications and an Air--Ground Scenario Graph (AGSG) grounded in simulator measurements. Deterministic modules then resolve spatial and temporal feasibility, while a managed executor synchronizes the ground and aerial simulators and evaluates trace-level requirements. When a violation occurs, the AGSG localizes the implicated components and constrains the repair scope. The language model is therefore used for semantic interpretation and guarded edit proposals, while simulator-bound quantities are grounded and validated through deterministic procedures.

\begin{figure*}[!t]
\centering
\includegraphics[width=0.96\textwidth]{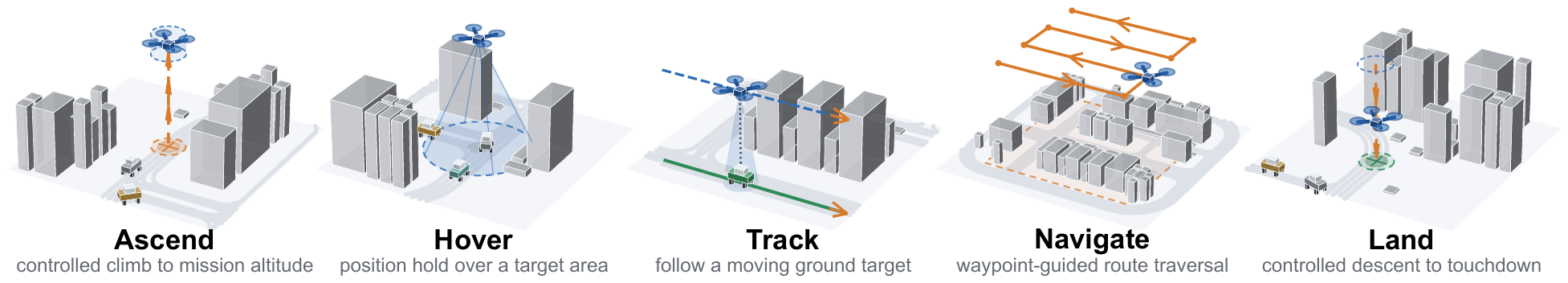}
\caption{UAV mission primitives in AURORA: ascend, hover, track, navigate, and land.}
\label{fig:tasks}
\end{figure*}

AURORA supports three core UAV mission modes---hover, track, and waypoint navigation---each executed between controlled ascent and landing phases (Fig.~\ref{fig:tasks}). Together, these primitives cover stationary aerial observation, dynamic air--ground following, and route-based missions while sharing a common execution and verification structure. 

The key distinction is between \emph{execution} and \emph{realization}. AURORA does not treat simulator completion as success; it verifies whether the requested relations and outcomes hold in the execution trace. The AGSG links each requirement to its dependencies, verification conditions, and adjustable parameters, enabling unmet requirements to be localized and repaired without regenerating the full scenario. Verification thus guides both scenario generation and refinement rather than serving as a post hoc check.

The contributions are three-fold:

\begin{enumerate}

\item \textbf{AURORA for verified air--ground scenario generation.} We formulate natural-language-driven air--ground scenario generation as a verified compilation process and develop an end-to-end agentic workflow that integrates simulator-grounded understanding, joint spatial and temporal reasoning, synchronized co-simulation, runtime verification, and bounded repair.

\item \textbf{Air--Ground Scenario Graph for verification and repair.} We introduce the AGSG, a typed intermediate representation that links requested relations to executable verification conditions, execution evidence, and the parameters that can affect them. This shared representation supports pre-execution feasibility checking, trace-based monitoring, failure localization, and localized repair.

\item \textbf{AURORA-Bench for evaluating scenario realization.} We construct a dedicated benchmark for language-generated air--ground co-simulation, comprising 200 prompts, 904 annotated requirements, and boundary-feasible and infeasible cases. A six-configuration, five-LLM evaluation separates execution reliability, prompt fidelity, and runtime realization, revealing failures that completion-based evaluation would otherwise overlook.

\end{enumerate}

\section{Related Work}
\label{sec:related}

\subsection{Air--ground simulation and cooperative perception}
Aerial sensing complements ground-based perception by reducing occlusion and extending coverage. Drone-assisted datasets demonstrate the value of vehicle--UAV collaboration across diverse environments~\cite{gao2025airv2x,wang2026griffin,ye2024uav3d,cui2026airsimag}, extending prior vehicle--vehicle and vehicle--infrastructure cooperative perception and reasoning~\cite{xu2021opv2v,yu2022dair,you2026v2x}. CARLA and AirSim provide widely used ground and aerial simulation capabilities~\cite{dosovitskiy2017carla,shah2017airsim}, while newer platforms integrate multirotor and vehicle dynamics~\cite{zeng2026carla}; OpenCDA supports cooperative driving automation~\cite{xu2021opencda}. Despite these capabilities, scenario construction and cross-domain coordination remain largely manual---valid placement, UAV mission design, synchronization, event coordination---and AURORA targets this authoring and verification layer rather than introducing another simulator.

\subsection{Scenario specification and language-driven generation}
Scenic~\cite{fremont2019scenic} and OpenSCENARIO~\cite{asam2022openscenario} provide structured traffic-scenario representations, but require specialized languages or predefined schemas. Learning-based methods further automate traffic generation and planning, including trajectory-generation approaches~\cite{zhong2022guided,wu2025hypergraph,gan2025planning}. Recent language-driven methods, including LLMScenario~\cite{chang2024llmscenario}, TTSG~\cite{ruan2024traffic}, LCTGen~\cite{tan2023language}, ChatScene~\cite{zhang2024chatscene}, and Talk2Traffic~\cite{sheng2025talk2traffic}, generate or interactively edit traffic scenarios from high-level descriptions, while ScenarioNet~\cite{NEURIPS2023_0c26a501} provides a data-driven framework for large-scale scenario modeling and generation. These approaches reduce manual specification but primarily target ground traffic and scene or trajectory generation. Air--ground scenarios additionally couple road geometry, three-dimensional flight, sensing, communication, and event timing, requiring a representation that captures cross-domain dependencies and joint realizability.

\begin{figure*}[!t]
  \centering
  \includegraphics[width=\textwidth]{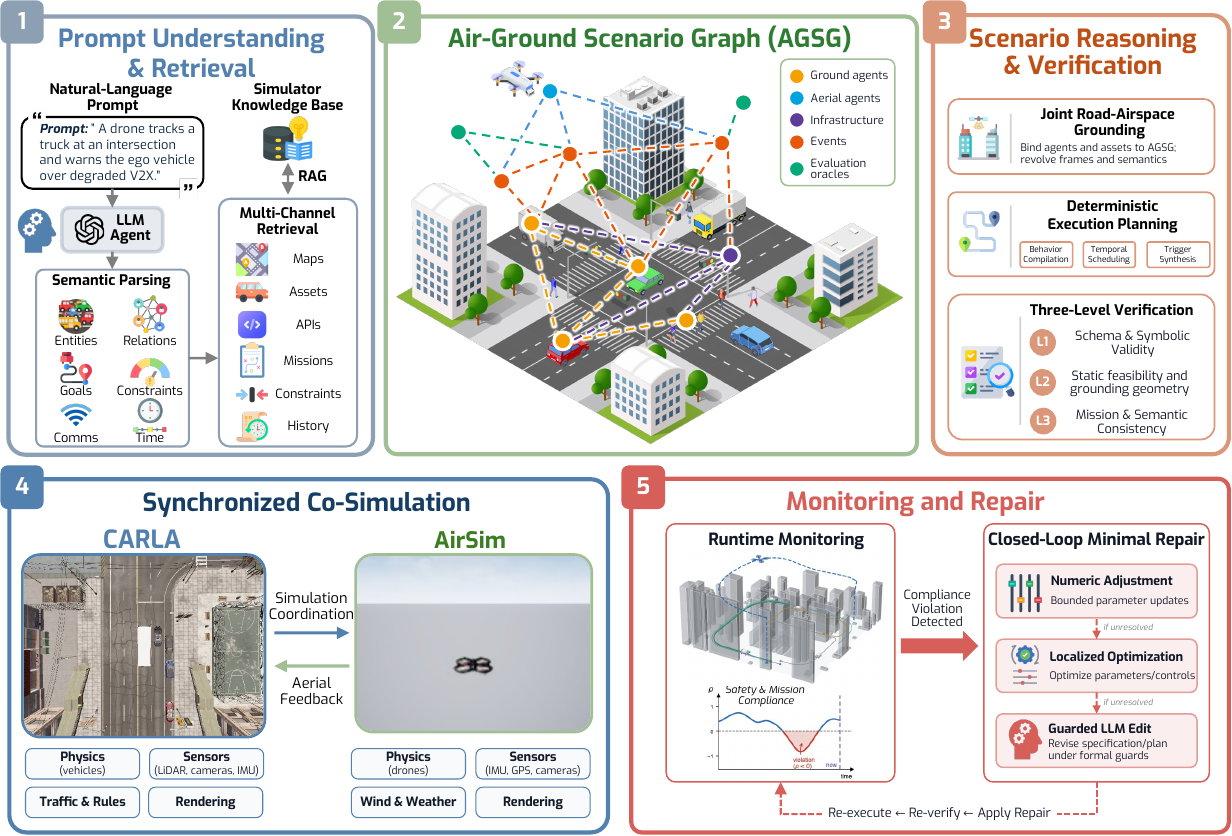}
  \caption{Overview of AURORA. Retrieval-grounded parsing constructs the AGSG for joint road--airspace grounding, temporal planning, and feasibility checks. Synchronized CARLA--AirSim execution provides requirement-level measurements for graph-based failure localization and bounded repair.}
  \label{fig:overview}
\end{figure*}

\subsection{Language-model execution and scenario verification}
Language models have also been explored for executable code, embodied control, and transportation reasoning~\cite{chen2021evaluating,liang2023code,yao2022react,wu2025v2x,gan2025planning}, and have recently been applied to natural-language and agentic UAV control in simulated environments~\cite{phadke2024integrating,shibu2026skysim,koubaa2025agentic}. Closed-loop UAV systems have further used execution feedback to evaluate and refine LLM-generated operation code~\cite{wang2025large}. However, successful execution does not guarantee that a generated scenario satisfies the request, and simulator scripts typically lack an explicit dependency structure for detecting and localizing semantic failures. VerifAI~\cite{dreossi2019verifai} demonstrates specification-based simulation analysis, while SafeBench~\cite{xu2022safebench} provides systematic benchmarking of autonomous-driving safety. These approaches, however, do not address natural-language compilation and verification of coupled air--ground scenarios. AURORA bridges these directions through an Air--Ground Scenario Graph that supports feasibility checking, runtime verification, failure tracing, and localized repair.

\section{Methods}
\label{sec:methods}

\subsection{Overview and Problem Formulation}
\label{sec:overview}

AURORA converts a natural-language request into an executable and verified air--ground co-simulation. The framework receives a prompt $p$, a simulation environment $\mathcal{E}$ specifying the simulator backends and map, and a knowledge base $\mathcal{K}$ constructed offline from direct measurements of $\mathcal{E}$. The objective is to generate a scenario whose execution realizes the spatial, temporal, behavioral, and communication requirements expressed in the prompt. Executability and requirement satisfaction are treated separately: a scenario may complete without runtime errors while failing to realize the requested interaction.

AURORA represents the request through a typed scenario specification:
\begin{equation}
\label{eq:specification}
\mathcal{S}
=
(\mathcal{W},\mathcal{A},u,\mathcal{M},\mathcal{V}_{E},\mathcal{O}),
\end{equation}
comprising world settings $\mathcal{W}$, ground agents and behaviors $\mathcal{A}$, UAV mission $u$, communication links $\mathcal{M}$, trigger--action events $\mathcal{V}_{E}$, and executable success conditions $\mathcal{O}$. The specification separates discrete scenario structure from the quantitative parameters that instantiate it, such as positions, altitudes, distances, timing thresholds, and communication settings. Each parameter also retains provenance indicating whether it is \emph{explicit}, \emph{inferred}, or \emph{default}. This distinction allows later repair to preserve user-specified values more strongly while giving greater flexibility to values introduced by the parser or schema.

The generation and verification process is formulated as
\begin{equation}
\label{eq:mapping}
(\mathcal{S}^{*},\mathcal{H}^{*},R^{*})
=
\operatorname{AURORA}(p;\mathcal{E},\mathcal{K}),
\end{equation}
where $\mathcal{S}^{*}$ is the returned specification, $\mathcal{H}^{*}$ its execution trace, and $R^{*}$ the verification and repair report. A successful output must pass pre-execution feasibility checks and satisfy all encoded success conditions during execution. Rather than returning only a completed simulation, AURORA retains the evidence needed to interpret the outcome, including violated conditions, their satisfaction margins, and any repairs applied.

AURORA follows three phases (Fig.~\ref{fig:overview}). During \emph{understanding}, retrieval-grounded parsing converts the prompt into a typed specification and its Air--Ground Scenario Graph (AGSG). During \emph{reasoning}, deterministic modules determine where and when the requested relations can be realized and whether the grounded configuration is feasible before execution. During \emph{orchestration}, a managed executor synchronizes the simulators, records the resulting trace, and evaluates trace-level requirements. Violations are mapped through the AGSG to a constrained set of candidate edits, allowing verification feedback to directly guide subsequent repair. The LLM therefore handles semantic interpretation and guarded symbolic proposals, while simulator-bound grounding, validation, execution, and numerical evaluation remain deterministic.

\subsection{Prompt Understanding and Scenario Representation}
\label{sec:representation}

The LLM agent interprets the prompt using the scenario schema and retrieved simulator knowledge, producing a typed specification rather than simulator code. The schema constrains admissible agent types, mission modes, event structures, parameters, units, and references. Invalid outputs are returned to the agent with explicit feedback from schema and static checks, allowing the interpretation to be revised through bounded retries. Accepted fields retain their provenance, supporting both prompt-fidelity evaluation and later control over which fields may be modified during repair.

Retrieval draws on a knowledge base $\Kb$ built offline from direct simulator measurements and past executions. It contains verified spawn locations and map attributes, measured asset geometry, UAV mission templates and operating limits, feasibility constraints, simulator interfaces, and execution history. Core registries remain available throughout generation, while retrieval selects prompt-relevant assets, spawn examples from the active map, and prior episodes with similar requirements. This measurement-grounded knowledge reduces dependence on potentially inaccurate model knowledge of simulator APIs or geometry.

Importantly, the parser and verifier share the same simulator-grounded records. The knowledge base therefore acts as a common source of constraints: quantities presented to the LLM during interpretation are drawn from the same environment knowledge later used to evaluate feasibility. This alignment reduces inconsistencies in which a value appears plausible during generation but violates the actual simulator configuration during execution.

The validated specification is converted deterministically into an \emph{Air--Ground Scenario Graph} (AGSG):
\begin{equation}
\label{eq:agsg}
\mathcal{G}
=
\left(
\mathcal{V},
\mathcal{E}_{G},
\tau_{\mathcal{V}},
\tau_{\mathcal{E}},
\Theta,
\omega
\right),
\end{equation}
where $\mathcal{V}$ contains environment, ground-agent, UAV, communication, event, and success-condition nodes; $\mathcal{E}_{G}$ encodes their relations and dependencies; $\tau_{\mathcal{V}}$ and $\tau_{\mathcal{E}}$ assign node and edge types; $\Theta$ contains quantitative parameters; and $\omega$ maps each parameter to its owning node. The resulting graph explicitly represents dependencies among scenario components---for example, placement affects sensing geometry, sensing may trigger communication, and communication may enable a subsequent vehicle action.

\begin{figure}[!t]
\centering
\includegraphics[width=\linewidth]{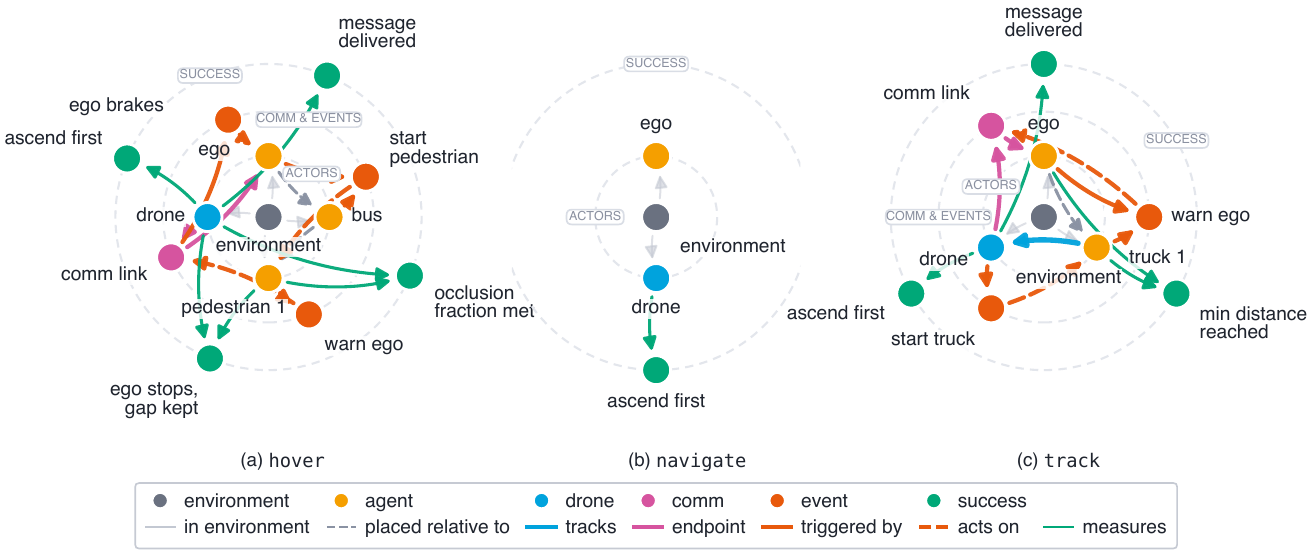}
\caption{Representative grounded AGSGs: (a) cooperative occlusion warning; (b) survey patrol; (c) tracking under degraded communication.}
\label{fig:agsg-examples}
\end{figure}

Rather than serving only as a scenario description, the AGSG links requested relations to their verification conditions and influencing parameters. Typed dependencies across spatial, temporal, perceptual, communication, and behavioral requirements allow the same graph to support feasibility checking, trace-based monitoring, failure localization, and repair. When a condition fails, its dependencies can be traced to the components and parameters that may have caused the violation. The AGSG therefore connects what the user requests, what the simulator executes, and what can be changed during repair. Fig.~\ref{fig:agsg-examples} shows representative grounded AGSGs across the supported mission modes.

\subsection{Joint Grounding, Planning, and Verification}
\label{sec:reasoning}

The agentic workflow delegates spatial grounding, temporal planning, and feasibility checking to deterministic modules. These modules determine where and when the scenario can unfold and return constraint violations to guide refinement. Together, they translate the relational specification into a configuration that is grounded in the actual road network, airspace, and execution constraints.

\paragraph{Joint road--airspace grounding}
A feasible location must simultaneously support the requested traffic event, UAV mission, sensing geometry, and communication geometry. Candidate road anchors are therefore ranked jointly as
\begin{equation}
\label{eq:joint}
s_{\mathrm{joint}}(r_i)
=
\alpha s_{\mathrm{road}}(r_i)
+\beta s_{\mathrm{air}}(r_i)
+\gamma s_{\mathrm{sense}}(r_i)
+\eta s_{\mathrm{comm}}(r_i),
\end{equation}
where the four terms respectively evaluate road suitability, aerial clearance, air-to-ground visibility, and communication geometry. Equal weights are used in the present implementation. The road term characterizes whether the local roadway can support the requested ground interaction, while the remaining terms assess whether the same location provides suitable airspace and air--ground relationships.

Once an anchor is selected, ground agents are placed on valid road or pedestrian elements using relative-placement rules. Requested placements that are not directly realizable are projected onto feasible lanes or pedestrian elements while preserving their intended spatial relations as closely as possible. UAV routes are then checked against an obstacle-height lattice, and infeasible waypoints or route segments are corrected before execution. This division keeps structural geometric correction within grounding rather than deferring such defects to the later runtime repair loop.

\paragraph{Temporal planning}
Each UAV mission is executed as a predicate-guarded state machine consisting of ascent, one of three supported mission modes (i.e., hover, track, or waypoint navigation), and landing (Fig.~\ref{fig:tasks}). Tasks are scheduled to respect physical feasibility and execution dependencies; sequential actions are completed in the required order before subsequent actions begin. For instance, navigation begins only after the UAV reaches its target altitude. Using realized predicates rather than fixed delays prevents downstream actions from beginning before their physical prerequisites are satisfied.

Inter-event timing is modeled using a Simple Temporal Network~\cite{dechter1991temporal}, where event orderings and deadlines are expressed as interval constraints. For events $e_i$ and $e_j$, temporal relations can be represented as $l_{ij} \leq t_{e_j}-t_{e_i} \leq u_{ij}$,
allowing requirements such as message transmission preceding reception or a warning arriving before a subsequent vehicle action to be checked before execution.

After grounding, geometry-dependent triggers are adjusted to ensure that the requested conditions remain reachable from the realized initial configuration. This step is important because a trigger defined against an imagined configuration may become immediately true or unreachable once agents are projected onto valid simulator geometry. Temporal planning therefore operates on the grounded scenario rather than independently of it.

\paragraph{Three-level verification}
Verification proceeds from inexpensive structural checks to trajectory-level requirements. First, \emph{schema verification} checks agent and task types, ranges, units, references, and temporal consistency. These checks identify malformed or internally inconsistent specifications before simulator execution. Second, \emph{static feasibility} checks whether placements, routes, relational constraints, and communication requirements can be realized in the grounded environment. These checks use the instantiated geometry and simulator-grounded limits rather than language-level plausibility alone. Violations are associated with the relevant fields or relations so that infeasible configurations can be revised before execution. Third, \emph{semantic verification} evaluates trace-dependent requirements such as visibility, message delivery, event ordering, and mission completion. These properties depend on interactions over time and therefore cannot be established from the initial configuration alone. They are monitored using quantitative temporal-logic semantics~\cite{maler2004monitoring,fainekos2009robustness,donze2010robust}, producing a signed margin $\rho_i$: $\rho_i\geq0$ indicates satisfaction, whereas $\rho_i<0$ quantifies the degree of violation. Unlike a binary pass/fail result, the signed margin also indicates how far a condition lies from satisfaction and can therefore guide subsequent correction. The same representation thus supports both verification and repair.

\begin{figure*}[!t]
\centering
\includegraphics[width=\textwidth]{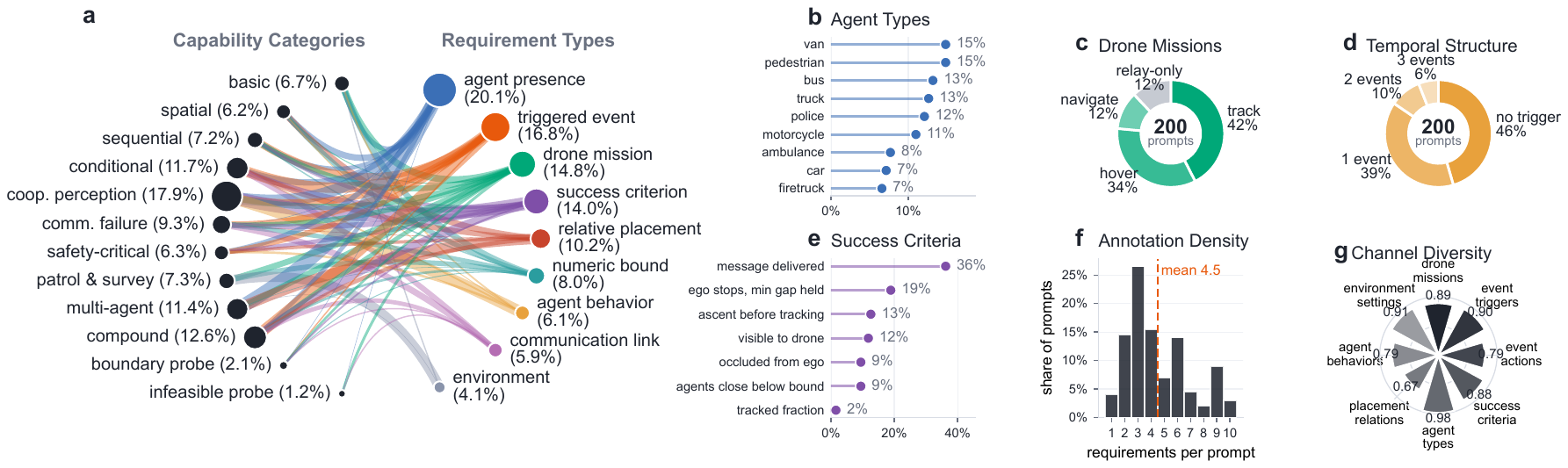}
\caption{AURORA-Bench composition, requirement annotations, and scenario diversity.}
\label{fig:bench}
\end{figure*}

\subsection{Synchronized Execution, Monitoring, and Repair}
\label{sec:execution}

\paragraph{Managed execution and synchronization}
A specification that passes pre-execution checks configures a fixed \emph{managed executor} for initialization, spawning, control, event scheduling, communication, logging, and cleanup. The LLM defines the scenario specification, while a validated execution path translates that specification into simulator operations. This separation avoids generating new simulator code for each scenario.

CARLA and AirSim advance under a shared logical clock. Coordinate transformation and a proxy UAV maintain a common ground--air representation, while online clock correction limits temporal drift. Communication is evaluated from realized sender--receiver geometry together with configured loss, latency, occlusion, and retransmission constraints. Agent states, message delivery, and event outcomes are recorded in the execution trace and evaluated against the encoded success conditions.

\paragraph{Runtime monitoring}
Pre-execution feasibility does not guarantee that the requested interaction will occur. Moving agents may fail to establish the intended relation, tracking may not converge, or an event may miss its timing or communication requirement. Runtime monitoring therefore evaluates success conditions on the realized trace and returns their signed robustness margins. This separates verified realization from nominal completion and quantifies how far each violated condition lies from satisfaction.

\paragraph{Bounded local repair}
Because trace-level verification requires co-simulation, AURORA uses bounded local repair:
\begin{equation}
\label{eq:reach}
\begin{aligned}
\mathcal{S}_0
&=
\operatorname{Ground}
\big(\operatorname{Parse}_{\mathcal{K}}(p)\big),\\
\mathcal{S}_{k+1}
&=
\operatorname{Repair}(\mathcal{S}_k,\mathcal{VS}_k)
\in\mathcal{N}(\mathcal{S}_k), \quad k<B,
\end{aligned}
\end{equation}
where $\mathcal{VS}_k=\{m:\rho_m(\mathcal{H}_k)<0\}$ identifies conditions violated by trace $\mathcal{H}_k$, and $B$ denotes the maximum number of repair iterations. The AGSG traces each violation to implicated parameters and fields, defining the admissible edit neighborhood $\mathcal{N}(\mathcal{S}_k)$. This localization prevents unrelated parts of the scenario from changing. Each numerical repair adjusts at most five parameters, while each symbolic repair revises at most six fields and must pass schema and static verification.

For coupled numerical failures, a kinematic replay surrogate evaluates candidate edits through a provenance-regularized objective:
\begin{equation}
\label{eq:repair}
\begin{split}
\Delta\Theta^{*}
=
\operatorname*{argmin}_{\Delta\Theta\in\Omega}\;&
\max\!\left(
0,\,
\varepsilon^{\star}
-
\min_m\hat{\rho}_m(\Theta+\Delta\Theta)
\right)
\\
&+
\lambda_{\mathrm{sem}}
\sum_j
w(c_j)
\frac{|\Delta\theta_j|}
{\operatorname{width}(\mathcal{D}_j)}.
\end{split}
\end{equation}
Here, $\Omega$ restricts edits to the localized parameter set and schema domains $\mathcal{D}_j$; $\hat{\rho}_m$ is the surrogate-estimated satisfaction margin; $c_j$ denotes the provenance class of parameter $\theta_j$; and $\varepsilon^{\star}>0$ specifies the desired robustness headroom. The first term drives the worst monitored condition toward a positive margin, while the second penalizes deviation from the parsed request after normalizing edits by their admissible ranges. The coefficient $\lambda_{\mathrm{sem}}$ balances satisfaction against semantic deviation, and $w(c_j)$ penalizes changes to explicit values more heavily than changes to inferred values or defaults.

The surrogate screens candidate edits before another full co-simulation run. Repair proceeds from single-parameter adjustment to localized multivariate search, followed by guarded LLM revision when numerical search cannot remove the violation. The LLM receives localized diagnostic feedback and may edit only within the admissible scope; each proposal must pass deterministic checks before re-execution. Structural geometric failures requiring broader changes, such as an infeasible route, are returned to the grounding stage rather than forced through the minimal-edit repair loop.

The loop terminates on success or budget exhaustion. The final report retains verification status, repair history, and unresolved margins. Because provenance weighting discourages but does not prohibit changes to explicit values, the report also distinguishes improvements in realized behavior from changes to the monitored conditions themselves.

\section{Experimental Design}
\label{sec:experiments}

\subsection{Benchmark and Evaluation Protocol}
\label{sec:benchmark}

AURORA-Bench comprises 200 prompts spanning 12 capability categories and 904 annotated requirements (Fig.~\ref{fig:bench}). Annotations include reference specifications, feasible realizations where applicable, variation ranges, and paraphrase groups. For prompts without a flight-task requirement, any feasible UAV mode is accepted. End-to-end evaluation uses a 50-prompt subset: 40 behavioral scenarios across nine categories, five boundary-feasible probes, and five infeasible probes.

We evaluate five LLMs---GPT-4o, GPT-5.5, GPT-5.4-mini, Gemini~3.1~Pro, and Gemini~3.8~Flash---using provider-default sampling with bridged CARLA~0.9.15 and AirSim~1.8.1. All configurations use the same prompts and simulation seed. Within each campaign, one model handles parsing, baseline code generation, and repair proposals. Retrieval history is frozen before each campaign to prevent information transfer from evaluation runs. Specifications, execution traces, verification reports, and model transcripts are archived. GPT-4o serves as the reference model for the illustrative example and aggregate analysis.

\subsection{Metrics and Baselines}
\label{sec:metrics}

We distinguish three properties of generated scenarios: \emph{executability}, \emph{prompt fidelity}, and \emph{runtime realization}. Executability asks whether a scenario completes without setup errors, runtime errors, or timeouts. Prompt fidelity measures whether the generated specification preserves the annotated request. Runtime realization asks whether the monitored conditions actually hold during execution. Accordingly, \emph{completion rate} (C.) measures successful execution, while \emph{runtime-failure rate} (RF) captures execution failures. \emph{Verified-pass rate} (P) measures executions in which all monitored conditions are satisfied, whereas \emph{silent-failure rate} (SF) captures completed runs with unmet requirements. \emph{Prompt fidelity} (Fid.) measures correspondence between the generated specification and annotated requirements.

Verification applies to the encoded monitored conditions, while fidelity and repair logs assess whether parsing and refinement preserve the original request, particularly when thresholds are modified. Supporting metrics characterize grounding accuracy, synchronization, event timing, communication, mission performance, and computational cost.

Six configurations progressively introduce framework components: direct code generation (\textbf{B1}), simulator documentation (\textbf{B2}), simulator-grounded retrieval (\textbf{B3}), the typed AGSG and managed executor without online verification (\textbf{B4}), static verification (\textbf{B5}), and runtime monitoring with localized repair (\textbf{AURORA}). Successful B1--B3 runs are classified as \emph{completed-unverifiable}, as they lack executable specifications for the semantic evaluation used here. Completion is compared across all six configurations, while verified outcomes are compared across B4, B5, and AURORA.

\section{Results}
\label{sec:results}

\subsection{Illustrative Example}
\label{sec:cases}

\begin{figure*}[!t]
\centering
\includegraphics[width=\textwidth]{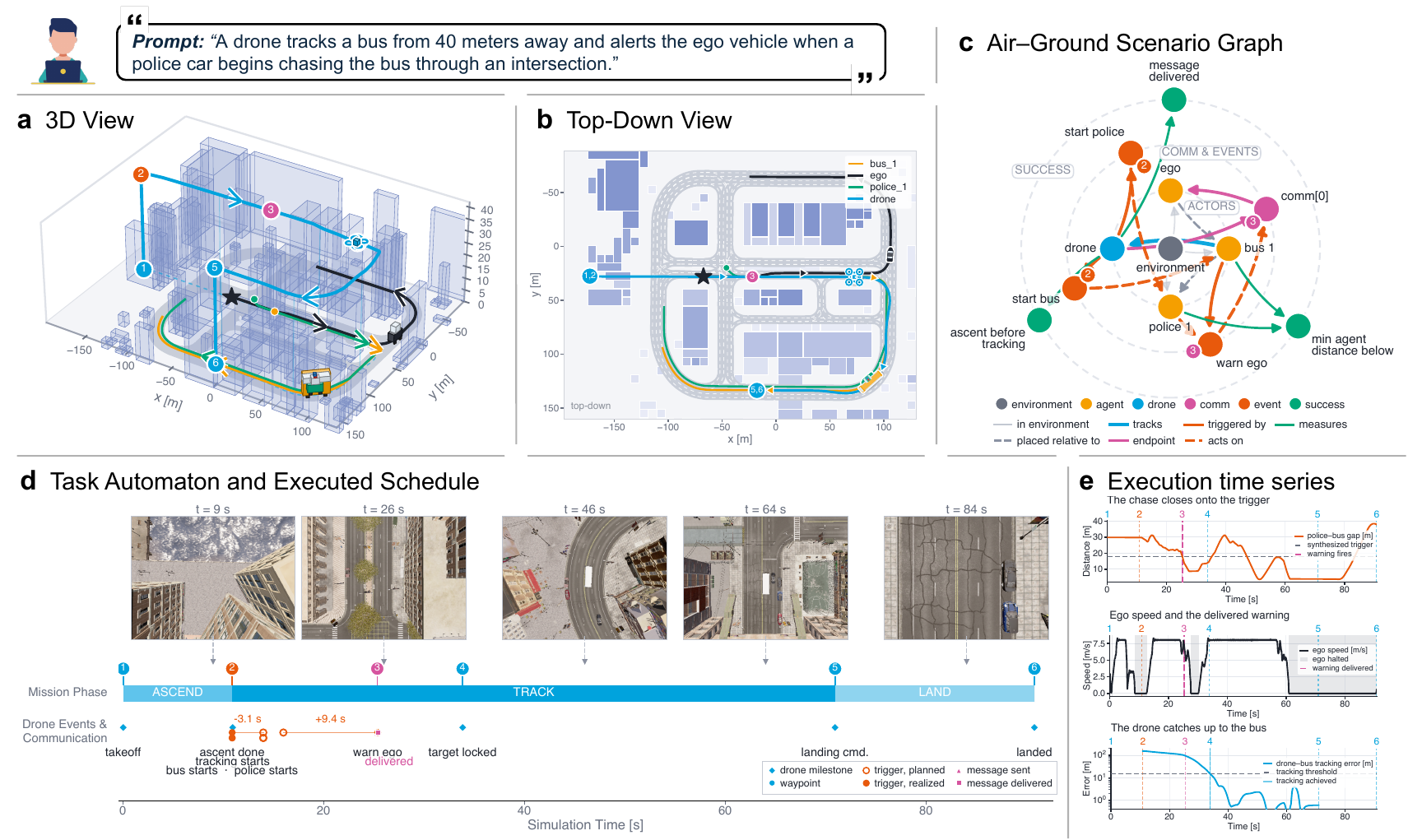}
\caption{Tracking scenario linking the language request, AGSG, grounded trajectories, and monitored outcomes.}
\label{fig:scenario-exp2}
\end{figure*}

Fig.~\ref{fig:scenario-exp2} illustrates how AURORA translates a tracking request into coordinated air--ground behavior. The UAV completes ascent before tracking the bus, while the police--bus interaction triggers a warning to the ego vehicle. The AGSG connects the participating agents, mission, communication link, and event dependencies to the grounded trajectories and execution schedule. These complementary views show not only how the requested interaction develops over time, but also whether its prerequisites and dependent events are realized in the intended order.

The warning reaches the ego vehicle before the UAV establishes close tracking. This sequence highlights why communication and mission performance require separate verification: successful message delivery does not establish sustained tracking. The event records confirm the warning outcome, while the UAV--bus distance trace reveals tracking convergence and subsequent variation. A single completion label would collapse these outcomes, whereas requirement-level monitoring preserves their distinction. By linking each condition to execution evidence, the AGSG also provides a basis for targeted diagnosis when part of the requested interaction remains unmet.

\begin{figure}[!t]
\centering
\includegraphics[width=\linewidth]{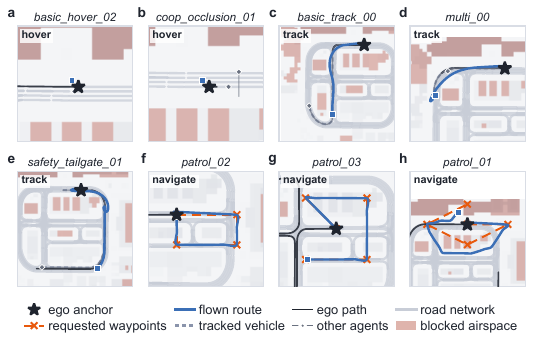}
\caption{Representative grounded hover, track, and navigation scenarios with UAV and ground-agent trajectories and obstacle-aware route correction.}
\label{fig:grounding-maps}
\end{figure}

Fig.~\ref{fig:grounding-maps} broadens the example across the three supported mission modes. Hover scenarios maintain aerial coverage around a grounded region, tracking scenarios couple UAV motion to a moving ground target, and navigation scenarios realize waypoint-based routes within the available airspace. Most requested routes are preserved directly, while geometrically infeasible waypoints are corrected during grounding before execution. Together with Fig.~\ref{fig:scenario-exp2}, these examples illustrate how AURORA translates different language-level mission structures into grounded and executable air--ground scenarios.

\begin{figure*}[!t]
\centering
\begin{subfigure}{\textwidth}
\centering
\includegraphics[width=0.9\textwidth]{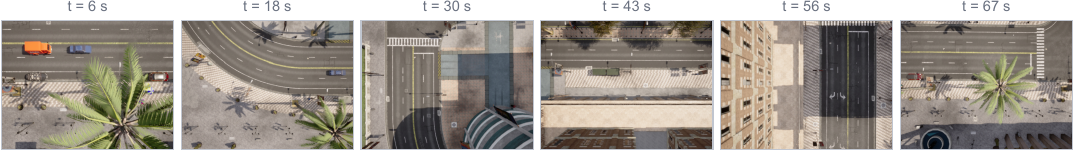}
\caption{``In clear weather at midday, a drone flies a rectangular survey route at 30 meters altitude over the streets around the ego car.''}
\label{fig:camera-patrol}
\end{subfigure}\\[4pt]
\begin{subfigure}{\textwidth}
\centering
\includegraphics[width=0.9\textwidth]{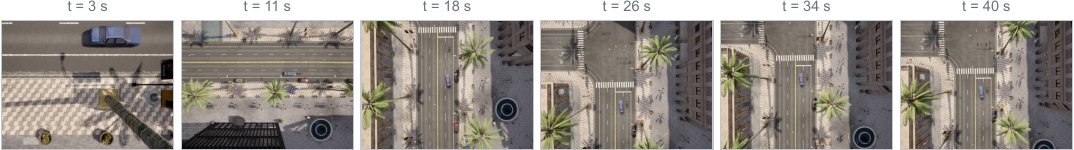}
\caption{``A drone ascends to 40 meters and then follows a motorcycle as it drives through the streets.''}
\label{fig:camera-basic-track}
\end{subfigure}\\[4pt]
\begin{subfigure}{\textwidth}
\centering
\includegraphics[width=0.9\textwidth]{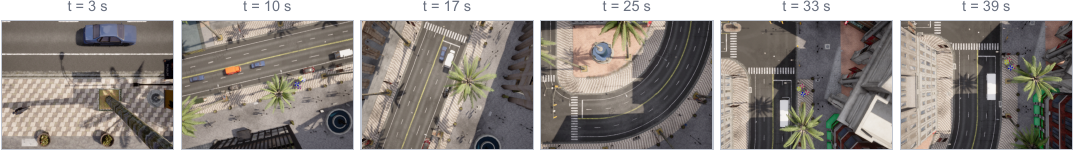}
\caption{``A bus drives ahead of the ego car and an ambulance follows behind it while two background cars circulate; a drone tracks the bus from 35 meters.''}
\label{fig:camera-multi}
\end{subfigure}\\[4pt]
\begin{subfigure}{\textwidth}
\centering
\includegraphics[width=0.9\textwidth]{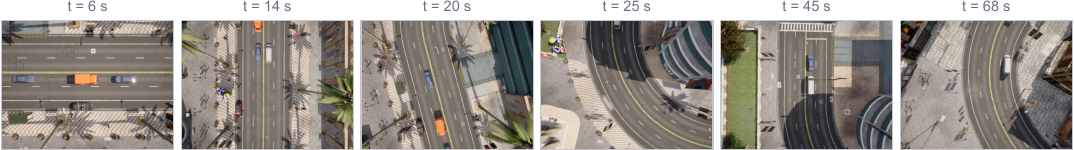}
\caption{``A van tailgates the ego car closely, closing from 45 meters; a tracking drone alerts the ego when the van comes within 20 meters.''}
\label{fig:camera-tailgate}
\end{subfigure}
\caption{Prompt-to-execution examples across four scenarios, showing six onboard UAV views sampled over each execution and the corresponding natural-language request.}
\label{fig:camera-grid}
\end{figure*}

Fig.~\ref{fig:camera-grid} shows the executed scenarios from the UAV camera, with one strip per request. The navigation example follows the requested survey route, while the tracking examples keep the motorcycle, bus, and tailgating van in view as the interactions unfold. These views provide qualitative evidence that the requested spatial and tracking interactions are realized in simulation, complementing the trace-based verification of communication and event outcomes.

\subsection{End-to-End Performance}
\label{sec:performance}

\begin{table*}[!t]
\centering
\caption{End-to-end performance across six configurations and five language models. Best value in each column is underlined.}
\label{tab:models}
\scriptsize
\setlength{\tabcolsep}{1.6pt}
\renewcommand{\arraystretch}{1.15}
\begin{tabular}{@{}lrrrrrrrrrrrrrrrrrrrrrrrrr@{}}
\toprule
& \multicolumn{5}{c}{GPT-4o} & \multicolumn{5}{c}{GPT-5.5} & \multicolumn{5}{c}{GPT-5.4-mini} & \multicolumn{5}{c}{Gemini 3.1 Pro} & \multicolumn{5}{c}{Gemini 3.8 Flash} \\
\cmidrule(lr){2-6}\cmidrule(lr){7-11}\cmidrule(lr){12-16}\cmidrule(lr){17-21}\cmidrule(lr){22-26}
Method & C.$\uparrow$ & RF$\downarrow$ & P$\uparrow$ & SF$\downarrow$ & Fid.$\uparrow$ & C.$\uparrow$ & RF$\downarrow$ & P$\uparrow$ & SF$\downarrow$ & Fid.$\uparrow$ & C.$\uparrow$ & RF$\downarrow$ & P$\uparrow$ & SF$\downarrow$ & Fid.$\uparrow$ & C.$\uparrow$ & RF$\downarrow$ & P$\uparrow$ & SF$\downarrow$ & Fid.$\uparrow$ & C.$\uparrow$ & RF$\downarrow$ & P$\uparrow$ & SF$\downarrow$ & Fid.$\uparrow$ \\
\midrule
\textbf{B1}: Direct Code Generation & 0.42 & 0.58 & --- & --- & --- & 0.82 & 0.18 & --- & --- & --- & 0.70 & 0.30 & --- & --- & --- & 0.80 & 0.20 & --- & --- & --- & 0.72 & 0.28 & --- & --- & --- \\
\textbf{B2}: $+$ Documentation & 0.68 & 0.32 & --- & --- & --- & 0.92 & 0.08 & --- & --- & --- & 0.86 & 0.14 & --- & --- & --- & 0.94 & 0.06 & --- & --- & --- & 0.92 & 0.08 & --- & --- & --- \\
\textbf{B3}: $+$ Grounded Retrieval & 0.66 & 0.34 & --- & --- & --- & 0.92 & 0.08 & --- & --- & --- & 0.78 & 0.22 & --- & --- & --- & 0.98 & 0.02 & --- & --- & --- & 0.98 & 0.02 & --- & --- & --- \\
\textbf{B4}: $+$ AGSG \& Executor & \underline{1.00} & \underline{0} & 0.58 & 0.42 & 0.966 & \underline{1.00} & \underline{0} & 0.42 & 0.58 & \underline{1.000} & \underline{1.00} & \underline{0} & 0.50 & 0.50 & \underline{1.000} & \underline{1.00} & \underline{0} & 0.66 & 0.34 & 0.993 & \underline{1.00} & \underline{0} & 0.62 & 0.38 & \underline{0.984} \\
\textbf{B5}: $+$ Static Verification & 0.98 & \underline{0} & 0.56 & 0.42 & \underline{0.980} & \underline{1.00} & \underline{0} & 0.48 & 0.52 & \underline{1.000} & \underline{1.00} & \underline{0} & 0.60 & 0.40 & 0.995 & 0.98 & \underline{0} & 0.70 & 0.28 & \underline{0.995} & \underline{1.00} & \underline{0} & 0.54 & 0.46 & 0.971 \\
\textbf{AURORA} & 0.98 & \underline{0} & \underline{0.76} & \underline{0.22} & \underline{0.980} & \underline{1.00} & \underline{0} & \underline{0.64} & \underline{0.36} & \underline{1.000} & \underline{1.00} & \underline{0} & \underline{0.70} & \underline{0.30} & 0.995 & 0.98 & \underline{0} & \underline{0.84} & \underline{0.14} & \underline{0.995} & \underline{1.00} & \underline{0} & \underline{0.74} & \underline{0.26} & 0.971 \\
\bottomrule
\end{tabular}
\end{table*}

\begin{figure}[!t]
\centering
\includegraphics[width=0.98\linewidth]{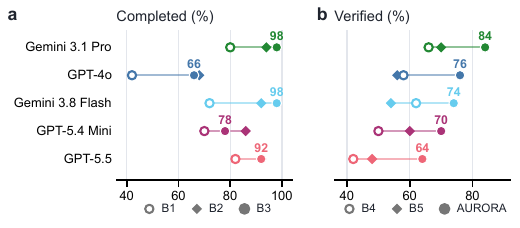}
\caption{Configuration comparison across models: (a) completion for code-generation baselines; (b) verified passes for AGSG-based configurations.}

\label{fig:ladder-models}
\end{figure}

The six configurations across five LLMs isolate the effects of simulator knowledge, structured execution, and verification-driven repair (Table~\ref{tab:models} and Fig.~\ref{fig:ladder-models}). The progression also separates improvements in executability from improvements in realized behavior, which do not necessarily arise from the same framework components.

\paragraph{Execution reliability}
Simulator knowledge improves code execution but does not eliminate runtime failures. Documentation raises completion across all five LLMs, while retrieval provides mixed additional gains, indicating that additional context alone does not guarantee reliable execution. The larger change comes from structured execution: introducing the typed AGSG and managed executor (B4) raises completion to 100\% across all models and eliminates observed runtime failures. Yet 34--58\% of scenarios remain silent failures. Where later configurations have slightly lower completion despite zero runtime failures, the difference reflects pre-execution rejection rather than execution breakdown. Reliable execution therefore does not imply successful realization, making completion alone insufficient for evaluating generated scenarios.

\begin{figure}[!t]
\centering
\includegraphics[width=0.92\linewidth]{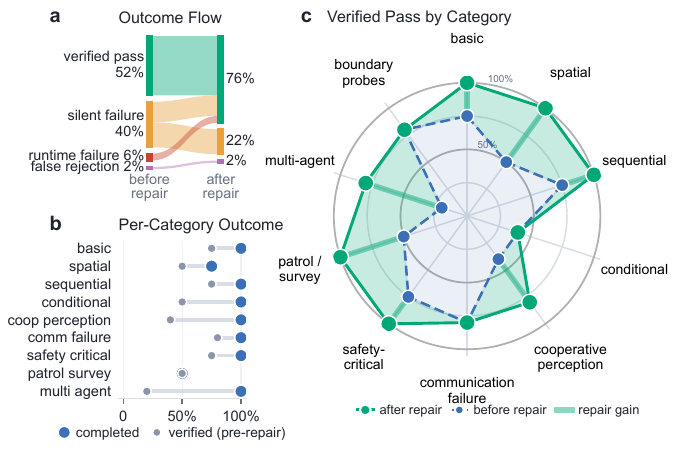}
\caption{GPT-4o evaluation: (a) repair outcomes; (b) completion and initial verification by category; (c) category-level repair gains.}

\label{fig:eval-overview}
\end{figure}

\paragraph{Verification and repair}
Runtime feedback provides the main gain in scenario realization. Static verification produces mixed changes in verified-pass rate, because identifying a feasible configuration before execution does not guarantee that its trace-dependent requirements will hold. Adding runtime monitoring and localized repair improves verified-pass rates by 10--20 percentage points over B5 across the five models, reaching 64--84\% under AURORA. The gains vary by scenario category (Fig.~\ref{fig:eval-overview}): spatial, sequential, and patrol scenarios improve substantially, while conditional scenarios remain more challenging. This pattern suggests that localized repair is most effective when a violation can be traced to a compact set of geometric, mission, or timing parameters; longer conditional chains remain harder to resolve. Remaining violations are reported explicitly rather than being absorbed into completion.

\paragraph{Prompt fidelity and realized behavior}
Prompt fidelity and runtime realization capture different properties. GPT-5.5, for example, achieves perfect specification fidelity without achieving the highest verified-pass rate. Preserving the request during parsing is therefore necessary but not sufficient; the grounded geometry, parameters, event timing, and mission dynamics must also realize the required conditions during execution. Notably, reported fidelity remains unchanged from B5 to AURORA for all five models, while verified-pass rates increase. The repair gains therefore do not coincide with a broad reduction in specification fidelity, although individual edits still require inspection when thresholds or explicit conditions are modified. Fidelity and verification thus provide complementary evidence of requirement preservation and realized behavior. Because B1--B3 lack executable specifications for semantic verification, their completion rates alone cannot establish whether the generated scenarios fulfill the requests.

\subsection{Grounding and Execution Validation}
\label{sec:validation}

\begin{figure}[!t]
\centering
\includegraphics[width=0.99\linewidth]{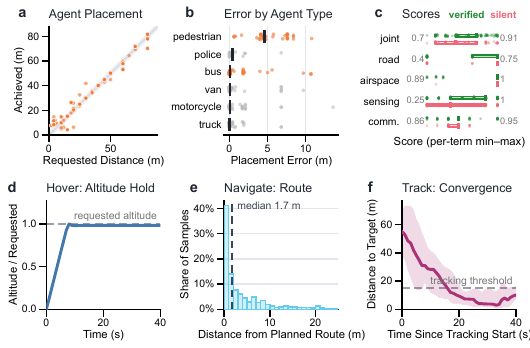}
\caption{Grounding validation: placement accuracy and per-type error (a,~b), anchor scores by outcome (c), and mission realization (d--f).}
\label{fig:analysis-grounding}
\end{figure}

We next examine scenario grounding and coordinated execution across the coupled simulators.

\paragraph{Spatial grounding and mission realization}
Fig.~\ref{fig:analysis-grounding} shows close agreement between requested and realized scenario geometry. Ground-agent placements are generally accurate, with larger deviations mainly caused by projection onto valid sidewalks or lanes. UAV missions likewise remain close to their intended geometry, with navigation deviations concentrated around turns. Most geometric differences therefore reflect feasibility enforcement rather than grounding failure. Tracking adds a temporal dimension: valid initialization does not ensure immediate mission fulfillment, because the UAV must first converge to the moving target and maintain the required relation over time. Placement accuracy and mission realization consequently answer different questions---whether the scenario is grounded correctly at initialization and whether the requested relation persists once the agents begin to move.

\paragraph{Grounding quality and scenario outcomes}
Verified runs have higher joint anchor scores than silent failures, with the largest separation in the sensing term (Fig.~\ref{fig:analysis-grounding}). Road, airspace, and communication scores differ less between the two groups. Successful grounding therefore depends not only on valid roads and aerial clearance, but also on whether the selected location supports the sensing relationship required by the interaction. Joint grounding captures these constraints before execution, while the resulting traces reveal whether the anticipated relationships persist as agents move. The anchor score is therefore useful as a pre-execution indicator of scenario quality, but it cannot replace trace-level verification of the realized interaction.

\begin{figure}[!t]
\centering
\includegraphics[width=0.9\linewidth]{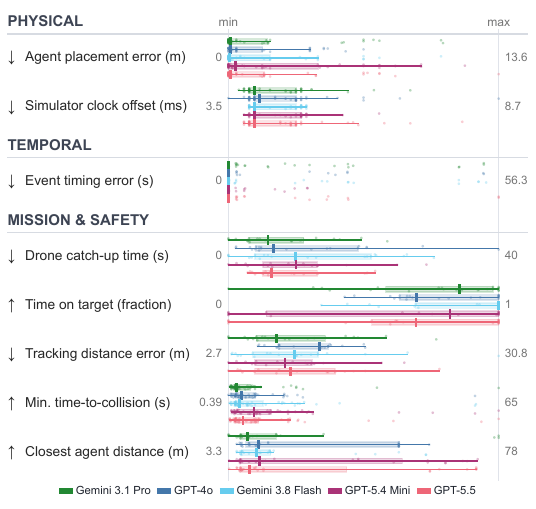}
\caption{Execution and mission metrics across language models.}
\label{fig:cap-models}
\end{figure}

\paragraph{Coordination across models}
Fig.~\ref{fig:cap-models} shows consistent simulator coordination across the five LLMs, with small clock offsets and limited event-timing error. The distributions nevertheless include delayed events and larger placement deviations, making variation beyond the median important. Mission performance varies more substantially, particularly in tracking convergence and time on target. Because all models use the same managed executor, this contrast separates low-level co-simulation reliability from model-dependent scenario realization. Stable synchronization provides the foundation for coordinated execution, but does not by itself guarantee the requested air--ground behavior.

\subsection{Repair Analysis and Computational Cost}
\label{sec:repair-cost}

\begin{figure}[!t]
\centering
\includegraphics[width=0.98\linewidth]{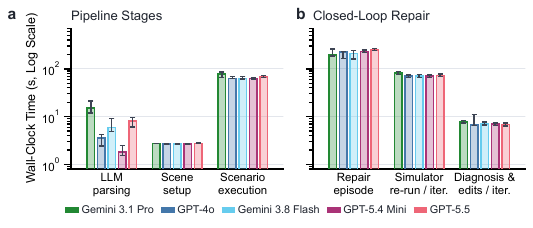}
\caption{Computational cost across models: (a) pipeline-stage latency; (b) repair-episode cost and per-iteration components.}
\label{fig:sys-models}
\end{figure}

We further analyze repair effectiveness and its computational cost.

\paragraph{Repair effectiveness}
Repair connects violated conditions to the components that can be adjusted. In a tracking example, two iterations move the tracked-fraction margin from negative to positive, while infeasible patrol routes are corrected during grounding through waypoint snapping. The two cases illustrate a useful division of responsibility: localized quantitative violations are handled by the repair loop, whereas structural geometric defects are returned to grounding rather than forced through small parameter edits. Routing failures to the appropriate stage avoids regenerating the full scenario while keeping each correction within its intended scope. Improved margins must nevertheless be interpreted alongside the edits that produced them. Some repairs relax acceptance thresholds, and parsing may reformulate an infeasible request instead of rejecting it. Provenance weighting discourages such changes but does not prohibit them. Repair logs therefore distinguish corrections to realized behavior from changes to the conditions used to evaluate it. Verified-pass rate should consequently be interpreted together with prompt fidelity and edit history rather than as a standalone measure of preservation of the original request.

\paragraph{Computational cost}
Fig.~\ref{fig:sys-models} shows that simulator execution dominates repair cost. Median repair episodes take 194--249\,s across models, largely due to repeated co-simulation runs, while diagnosis and edit generation contribute comparatively little. Parsing latency varies across LLMs, whereas static checks and kinematic preview remain inexpensive. This asymmetry supports placing inexpensive verification and surrogate screening before full re-execution whenever possible. Localized edits constrain the search space, but the larger opportunity for reducing repair cost lies in avoiding unnecessary simulator reruns rather than accelerating edit generation alone.

\section{Conclusion}
\label{sec:conclusion}

AURORA connects natural-language scenario generation with explicit feasibility checking, synchronized air--ground execution, and trace-based verification. Its Air--Ground Scenario Graph links requested relations to executable conditions, execution evidence, and repair variables, providing a shared representation for grounding, monitoring, failure localization, and refinement. Evaluation across five language models distinguishes three properties of generated scenarios: whether they execute, preserve the request, and realize the requested conditions. Structured execution improves reliability, but successful completion can still mask semantic failures; runtime verification exposes these failures, while localized repair resolves many without regenerating the full scenario. More broadly, the results support treating language-driven scenario generation as verified compilation rather than direct code generation. Explicit intermediate representations make cross-domain dependencies inspectable and provide a structured path from runtime violations to the parameters that can affect them, reducing reliance on language models for simulator-specific reasoning and improving failure diagnosis.

The current framework supports a defined scenario schema, three UAV mission templates, simplified sensing and communication models, and one simulation environment. Because parsing can reinterpret infeasible requests and repair can relax monitored thresholds, a verified pass should be interpreted as satisfaction of the encoded monitored conditions rather than unconditional satisfaction of the original prompt. Future work will strengthen preservation of explicit requirements and terminal handling of infeasible requests, while extending AURORA toward multi-UAV planning and control, richer cooperative air--ground interactions, vision--language--action models for closed-loop perception and decision making, and additional simulation environments and real-world testbeds.

\bibliographystyle{IEEEtran}
\bibliography{refs}

@inproceedings{dosovitskiy2017carla,
  title={CARLA: An open urban driving simulator},
  author={Dosovitskiy, Alexey and Ros, German and Codevilla, Felipe and Lopez, Antonio and Koltun, Vladlen},
  booktitle={Conference on robot learning},
  pages={1--16},
  year={2017},
  organization={PMLR}
}

@inproceedings{shah2017airsim,
  title={Airsim: High-fidelity visual and physical simulation for autonomous vehicles},
  author={Shah, Shital and Dey, Debadeepta and Lovett, Chris and Kapoor, Ashish},
  booktitle={Field and service robotics: Results of the 11th international conference},
  pages={621--635},
  year={2017},
  organization={Springer}
}

@article{ruan2024traffic,
  title={Traffic scene generation from natural language description for autonomous vehicles with large language model},
  author={Ruan, Bo-Kai and Tsui, Hao-Tang and Li, Yung-Hui and Shuai, Hong-Han},
  journal={arXiv preprint arXiv:2409.09575},
  year={2024}
}

@article{zeng2026carla,
  title={CARLA-Air: Fly Drones Inside a CARLA World--A Unified Infrastructure for Air-Ground Embodied Intelligence},
  author={Zeng, Tianle and Wen, Yanci and Zhang, Hong},
  journal={arXiv preprint arXiv:2603.28032},
  year={2026}
}

@article{gao2025airv2x,
  title={Airv2x: Unified air-ground vehicle-to-everything collaboration},
  author={Gao, Xiangbo and Wu, Yuheng and Yang, Fengze and Luo, Xuewen and Wu, Keshu and Chen, Xinghao and Wang, Yuping and Liu, Chenxi and Zhou, Yang and Tu, Zhengzhong},
  journal={arXiv preprint arXiv:2506.19283},
  year={2025}
}

@article{tan2023language,
  title={Language conditioned traffic generation},
  author={Tan, Shuhan and Ivanovic, Boris and Weng, Xinshuo and Pavone, Marco and Kraehenbuehl, Philipp},
  journal={arXiv preprint arXiv:2307.07947},
  year={2023}
}

@article{zhong2022guided,
  title={Guided conditional diffusion for controllable traffic simulation},
  author={Zhong, Ziyuan and Rempe, Davis and Xu, Danfei and Chen, Yuxiao and Veer, Sushant and Che, Tong and Ray, Baishakhi and Pavone, Marco},
  journal={arXiv preprint arXiv:2210.17366},
  year={2022}
}

@inproceedings{zhang2024chatscene,
  title={Chatscene: Knowledge-enabled safety-critical scenario generation for autonomous vehicles},
  author={Zhang, Jiawei and Xu, Chejian and Li, Bo},
  booktitle={Proceedings of the IEEE/CVF Conference on Computer Vision and Pattern Recognition},
  pages={15459--15469},
  year={2024}
}

@inproceedings{fremont2019scenic,
  title={Scenic: a language for scenario specification and scene generation},
  author={Fremont, Daniel J. and Dreossi, Tommaso and Ghosh, Shromona and Yue, Xiangyu and Sangiovanni-Vincentelli, Alberto L. and Seshia, Sanjit A.},
  booktitle={Proceedings of the 40th ACM SIGPLAN conference on programming language design and implementation},
  pages={63--78},
  year={2019}
}

@inproceedings{dreossi2019verifai,
  title={Verifai: A toolkit for the formal design and analysis of artificial intelligence-based systems},
  author={Dreossi, Tommaso and Fremont, Daniel J. and Ghosh, Shromona and Kim, Edward and Ravanbakhsh, Hadi and Vazquez-Chanlatte, Marcell and Seshia, Sanjit A.},
  booktitle={International Conference on Computer Aided Verification},
  pages={432--442},
  year={2019},
  organization={Springer}
}

@inproceedings{liang2023code,
  title={Code as policies: Language model programs for embodied control},
  author={Liang, Jacky and Huang, Wenlong and Xia, Fei and Xu, Peng and Hausman, Karol and Ichter, Brian and Florence, Pete and Zeng, Andy},
  booktitle={2023 IEEE International conference on robotics and automation (ICRA)},
  pages={9493--9500},
  year={2023},
  organization={IEEE}
}

@inproceedings{yao2022react,
  title={React: Synergizing reasoning and acting in language models},
  author={Yao, Shunyu and Zhao, Jeffrey and Yu, Dian and Shafran, Izhak and Narasimhan, Karthik R. and Cao, Yuan},
  booktitle={NeurIPS 2022 Foundation Models for Decision Making Workshop},
  year={2022}
}

@article{chen2021evaluating,
  title={Evaluating large language models trained on code},
  author={Chen, Mark and Tworek, Jerry and Jun, Heewoo and Yuan, Qiming and Pinto, Henrique Ponde De Oliveira and Kaplan, Jared and Edwards, Harri and Burda, Yuri and Joseph, Nicholas and Brockman, Greg and others},
  journal={arXiv preprint arXiv:2107.03374},
  year={2021}
}

@article{xu2021opv2v,
  title={Opv2v: An open benchmark dataset and fusion pipeline for perception with vehicle-to-vehicle communication},
  author={Xu, Runsheng and Xiang, Hao and Xia, Xin and Han, Xu and Li, Jinlong and Ma, Jiaqi},
  journal={arXiv preprint arXiv:2109.07644},
  year={2021}
}

@inproceedings{yu2022dair,
  title={Dair-v2x: A large-scale dataset for vehicle-infrastructure cooperative 3d object detection},
  author={Yu, Haibao and Luo, Yizhen and Shu, Mao and Huo, Yiyi and Yang, Zebang and Shi, Yifeng and Guo, Zhenglong and Li, Hanyu and Hu, Xing and Yuan, Jirui and others},
  booktitle={Proceedings of the IEEE/CVF conference on computer vision and pattern recognition},
  pages={21361--21370},
  year={2022}
}

@inproceedings{xu2021opencda,
  title={Opencda: an open cooperative driving automation framework integrated with co-simulation},
  author={Xu, Runsheng and Guo, Yi and Han, Xu and Xia, Xin and Xiang, Hao and Ma, Jiaqi},
  booktitle={2021 IEEE International Intelligent Transportation Systems Conference (ITSC)},
  pages={1155--1162},
  year={2021},
  organization={IEEE}
}

@inproceedings{maler2004monitoring,
  title={Monitoring temporal properties of continuous signals},
  author={Maler, Oded and Nickovic, Dejan},
  booktitle={International symposium on formal techniques in real-time and fault-tolerant systems},
  pages={152--166},
  year={2004},
  organization={Springer}
}

@article{dechter1991temporal,
  title={Temporal constraint networks},
  author={Dechter, Rina and Meiri, Itay and Pearl, Judea},
  journal={Artificial intelligence},
  volume={49},
  number={1-3},
  pages={61--95},
  year={1991},
  publisher={Elsevier}
}

@inproceedings{donze2010robust,
  title={Robust satisfaction of temporal logic over real-valued signals},
  author={Donz{\'e}, Alexandre and Maler, Oded},
  booktitle={International conference on formal modeling and analysis of timed systems},
  pages={92--106},
  year={2010},
  organization={Springer}
}

@article{fainekos2009robustness,
  title={Robustness of temporal logic specifications for continuous-time signals},
  author={Fainekos, Georgios E. and Pappas, George J.},
  journal={Theoretical Computer Science},
  volume={410},
  number={42},
  pages={4262--4291},
  year={2009},
  publisher={Elsevier}
}

@misc{asam2022openscenario,
  author       = {{ASAM e.V.}},
  title        = {{ASAM OpenSCENARIO}: Dynamic Content in Driving Simulation, Standard v1.2},
  howpublished = {\url{https://www.asam.net/standards/detail/openscenario/}},
  year         = {2022}
}

@inproceedings{wang2026griffin,
  title={Griffin: Aerial-ground cooperative detection and tracking dataset and benchmark},
  author={Wang, Jiahao and Cao, Xiangyu and Zhong, Jiaru and Zhang, Yuner and Han, Zeyu and Yu, Haibao and Zhang, Chuang and He, Lei and Xu, Shaobing and Wang, Jianqiang},
  booktitle={Proceedings of the AAAI Conference on Artificial Intelligence},
  volume={40},
  number={12},
  pages={9867--9875},
  year={2026}
}

@article{ye2024uav3d,
  title={Uav3d: A large-scale 3d perception benchmark for unmanned aerial vehicles},
  author={Ye, Hui and Sunderraman, Rajshekhar and Ji, Shihao},
  journal={Advances in Neural Information Processing Systems},
  volume={37},
  pages={55425--55442},
  year={2024}
}

@article{yang2026edge,
  title={Edge-based multimodal sensor data fusion with Vision-Language-Action (VLA) model for real-time autonomous vehicle accident avoidance},
  author={Yang, Fengze and Yu, Bo and Zhou, Yang and Luo, Xuewen and Tu, Zhengzhong and Liu, Chenxi},
  journal={Engineering Applications of Artificial Intelligence},
  volume={180},
  pages={115186},
  year={2026},
  publisher={Elsevier}
}

@inproceedings{feng2024u2udata,
  title={U2udata: A large-scale cooperative perception dataset for swarm uavs autonomous flight},
  author={Feng, Tongtong and Wang, Xin and Han, Feilin and Zhang, Leping and Zhu, Wenwu},
  booktitle={Proceedings of the 32nd ACM International Conference on Multimedia},
  pages={7600--7608},
  year={2024}
}

@inproceedings{gao2025langcoop,
  title={Langcoop: Collaborative driving with language},
  author={Gao, Xiangbo and Wu, Yuheng and Wang, Rujia and Liu, Chenxi and Zhou, Yang and Tu, Zhengzhong},
  booktitle={2025 IEEE/CVF Conference on Computer Vision and Pattern Recognition Workshops (CVPRW)},
  pages={4226--4237},
  year={2025},
  organization={IEEE}
}

@inproceedings{xu2022v2x,
  title={V2x-vit: Vehicle-to-everything cooperative perception with vision transformer},
  author={Xu, Runsheng and Xiang, Hao and Tu, Zhengzhong and Xia, Xin and Yang, Ming-Hsuan and Ma, Jiaqi},
  booktitle={European conference on computer vision},
  pages={107--124},
  year={2022},
  organization={Springer}
}

@article{cui2026airsimag,
  title={AirSimAG: A High-Fidelity Simulation Platform for Air-Ground Collaborative Robotics},
  author={Cui, Yangjie and Dong, Xin and Gao, Boyang and Xiang, Jinwu and Li, Daochun and Tu, Zhan},
  journal={arXiv preprint arXiv:2603.23079},
  year={2026}
}

@article{xu2022safebench,
  title={Safebench: A benchmarking platform for safety evaluation of autonomous vehicles},
  author={Xu, Chejian and Ding, Wenhao and Lyu, Weijie and Liu, Zuxin and Wang, Shuai and He, Yihan and Hu, Hanjiang and Zhao, Ding and Li, Bo},
  journal={Advances in Neural Information Processing Systems},
  volume={35},
  pages={25667--25682},
  year={2022}
}

@inproceedings{NEURIPS2023_0c26a501,
	author = {Li, Quanyi and Peng, Zhenghao (Mark) and Feng, Lan and Liu, Zhizheng and Duan, Chenda and Mo, Wenjie and Zhou, Bolei},
	booktitle = {Advances in Neural Information Processing Systems},
	doi = {10.52202/075280-0172},
	editor = {A. Oh and T. Naumann and A. Globerson and K. Saenko and M. Hardt and S. Levine},
	pages = {3894--3920},
	publisher = {Curran Associates, Inc.},
	title = {ScenarioNet: Open-Source Platform for Large-Scale Traffic Scenario Simulation and Modeling},
	url = {https://proceedings.neurips.cc/paper_files/paper/2023/file/0c26a501df8fb919a0350e2df06b5d39-Paper-Datasets_and_Benchmarks.pdf},
	volume = {36},
	year = {2023}}

@article{phadke2024integrating,
  title={Integrating large language models for uav control in simulated environments: A modular interaction approach},
  author={Phadke, Abhishek and Hadimlioglu, Alihan and Chu, Tianxing and Sekharan, Chandra N},
  journal={arXiv preprint arXiv:2410.17602},
  year={2024}
}

@article{shibu2026skysim,
  title={SkySim: A ROS2-based simulation environment for natural language control of drone swarms using large language models},
  author={Shibu, Aditya and Saleh, Marah and Al-Musleh, Mohamed and Abdulaziz, Nidhal},
  journal={arXiv preprint arXiv:2602.01226},
  year={2026}
}

@article{wang2025generative,
  title={Generative ai for autonomous driving: Frontiers and opportunities},
  author={Wang, Yuping and Xing, Shuo and Cui, Can and Li, Renjie and Hua, Hongyuan and Tian, Kexin and Mo, Zhaobin and Gao, Xiangbo and Pavone, Marco and Zhou, Yang and others},
  journal={ACM Computing Surveys},
  year={2025},
  publisher={ACM New York, NY}
}

@article{wu2025digital,
  title={A digital twin framework for physical-virtual integration in v2x-enabled connected vehicle corridors},
  author={Wu, Keshu and Li, Pei and Cheng, Yang and Parker, Steven T and Ran, Bin and Noyce, David A and Ye, Xinyue},
  journal={IEEE Transactions on Intelligent Transportation Systems},
  volume={26},
  number={9},
  pages={14407--14420},
  year={2025},
  publisher={IEEE}
}

@article{wu2025v2x,
  title={V2x-llm: Enhancing v2x integration and understanding in connected vehicle corridors},
  author={Wu, Keshu and Li, Pei and Zhou, Yang and Gan, Rui and You, Junwei and Cheng, Yang and Zhu, Jingwen and Parker, Steven T and Ran, Bin and Noyce, David A and others},
  journal={arXiv preprint arXiv:2503.02239},
  year={2025}
}

@article{wu2025hypergraph,
  title={Hypergraph-based motion generation with multi-modal interaction relational reasoning},
  author={Wu, Keshu and Zhou, Yang and Shi, Haotian and Lord, Dominique and Ran, Bin and Ye, Xinyue},
  journal={Transportation Research Part C: Emerging Technologies},
  volume={180},
  pages={105349},
  year={2025},
  publisher={Elsevier}
}

@article{zhang2025virtual,
  title={Virtual roads, smarter safety: A digital twin framework for mixed autonomous traffic safety analysis},
  author={Zhang, Hao and Yue, Ximin and Tian, Kexin and Li, Sixu and Wu, Keshu and Li, Zihao and Lord, Dominique and Zhou, Yang},
  journal={IEEE Internet of Things Journal},
  year={2025},
  publisher={IEEE}
}

@article{gan2025planning,
  title={Planning safety trajectories with dual-phase, physics-informed, and transportation knowledge-driven large language models},
  author={Gan, Rui and Li, Pei and Long, Keke and An, Bocheng and You, Junwei and Wu, Keshu and Ran, Bin},
  journal={arXiv preprint arXiv:2504.04562},
  year={2025}
}

@article{koubaa2025agentic,
  title={Agentic uavs: Llm-driven autonomy with integrated tool-calling and cognitive reasoning},
  author={Koubaa, Anis and Gabr, Khaled},
  journal={arXiv preprint arXiv:2509.13352},
  year={2025}
}

@article{wang2025large,
  title={Large language model-driven closed-loop uav operation with semantic observations},
  author={Wang, Wenhao and Li, Yanyan and Jiao, Long and Yuan, Jiawei},
  journal={IEEE Internet of Things Journal},
  year={2025},
  publisher={IEEE}
}

@inproceedings{sheng2025talk2traffic,
  title={Talk2Traffic: Interactive and Editable Traffic Scenario Generation for Autonomous Driving with Multimodal Large Language Model},
  author={Sheng, Zihao and Huang, Zilin and Qu, Yansong and Leng, Yue and Chen, Sikai},
  booktitle={Proceedings of the IEEE/CVF Conference on Computer Vision and Pattern Recognition (CVPR) Workshops},
  pages={3827--3836},
  year={2025}
}

@article{chang2024llmscenario,
  title={Llmscenario: Large language model driven scenario generation},
  author={Chang, Cheng and Wang, Siqi and Zhang, Jiawei and Ge, Jingwei and Li, Li},
  journal={IEEE Transactions on Systems, Man, and Cybernetics: Systems},
  volume={54},
  number={11},
  pages={6581--6594},
  year={2024},
  publisher={IEEE}
}

@article{gao2025automated,
  title={Automated vehicles should be connected with natural language},
  author={Gao, Xiangbo and Wu, Keshu and Zhang, Hao and Tian, Kexin and Zhou, Yang and Tu, Zhengzhong},
  journal={arXiv preprint arXiv:2507.01059},
  year={2025}
}

@article{you2026v2x,
  title={V2x-vlm: End-to-end v2x cooperative autonomous driving through large vision-language models},
  author={You, Junwei and Jiang, Zhuoyu and Huang, Zilin and Shi, Haotian and Gan, Rui and Wu, Keshu and Cheng, Xi and Li, Xiaopeng and Ran, Bin},
  journal={Transportation Research Part C: Emerging Technologies},
  volume={183},
  pages={105457},
  year={2026},
  publisher={Elsevier}
}

\end{document}